\documentclass[journal]{IEEEtran}
\usepackage{hyperref}
\usepackage[pdftex]{graphicx}
  \graphicspath{{../pdf/}{../jpeg/}}
	\DeclareGraphicsExtensions{.pdf,.jpeg,.png}
        
	\usepackage[cmex10]{amsmath}
	\usepackage{mathabx}
	\usepackage{algorithmic}
	\usepackage{array}
	\usepackage{mdwmath}

	\usepackage{mdwtab}
	\usepackage{eqparbox}
	\usepackage{url}
\newcommand{\specialcell}[2][c]{%
  \begin{tabular}[#1]{@{}c@{}}#2\end{tabular}}
\begin{document}
\title{MissBeamNet: Learning Missing Doppler Velocity Log Beam Measurements}


\author{\IEEEauthorblockN{Mor Yona and Itzik Klein}}
\maketitle

\begin{abstract}
One of the primary means of sea exploration is autonomous underwater vehicles (AUVs). To perform these tasks, AUVs must navigate the rough challenging sea environment. AUVs usually employ an inertial navigation system (INS), aided by a Doppler velocity log (DVL), to provide the required navigation accuracy. The DVL transmits four acoustic beams to the seafloor, and by measuring changes in the frequency of the returning beams, the DVL can estimate the AUV velocity vector. However, in practical scenarios, not all the beams are successfully reflected. When only three beams are available, the accuracy of the velocity vector is degraded. When fewer than three beams are reflected, the DVL cannot estimate the AUV velocity vector. This paper presents a data-driven approach, MissBeamNet, to regress the missing beams in partial DVL beam measurement cases. To that end, a deep neural network (DNN) model is designed to process the available beams along with past DVL measurements to regress the missing beams. The AUV velocity vector is estimated using the available measured and regressed beams. To validate the proposed approach, sea experiments were made with the "Snapir" AUV, resulting in an 11 hours dataset of DVL measurements. Our results show that the proposed system can accurately estimate velocity vectors in situations of missing beam measurements. Our dataset and codebase implementing the described framework is available
at our GitHub \href{https://github.com/ansfl/MissBeamNet}{repository}.
\end{abstract}

\IEEEoverridecommandlockouts

\IEEEpeerreviewmaketitle


\section{Introduction}\label{sec1}
The demand for autonomous underwater vehicles (AUV) is significantly growing \cite{Luo et al.2022,Mohammadi et al.2022,Wynn et al. 2014,Bovio et al. 2006}. AUVs are used in a variety of applications, such as seafloor exploration and mapping \cite{Kume et al. 2013}, pipeline inspection \cite{Niu et al. 2009,Hongwei et al. 2021}, and underwater mine detection \cite{Maussang et al. 2003}. An accurate navigation system is necessary for the AUV to navigate challenging sea conditions and successfully perform the required tasks. From a navigational perspective, the commonly used global navigation satellite system (GNSS) is unavailable underwater. Furthermore, underwater currents and the ever-changing landscape make it difficult to use simultaneous localization and mapping (SLAM) \cite{Palomer et al. 2016}. Consequently, most AUVs employ an inertial navigation system (INS) aided by a Doppler velocity log (DVL). The INS provides a complete navigation solution comprising position, velocity, and orientation using three-axis accelerometers and three-axis gyroscopes. However, due to inertial measurement errors, the pure inertial solution will drift over time \cite{Thong et al. 2002}. The DVL provides an accurate estimate of the AUV velocity vector, which is used to aid the INS and obtain an accurate navigation solution. The fusion between INS and DVL is well addressed in the literature under normal DVL operating conditions. For example, a rotational dynamic model was shown to improve the INS/DVL fusion performance \cite{Karmozdi et al. 2020}. Furthermore, an adaptive Kalman filter aimed at finding the optimal window length for each measurement has been suggested \cite{Emami and Taben. 2018}. In order to improve the extended Kalman filter, an innovative unscented Kalman filter was developed for AUV navigation \cite{Allotta et al. 2016}. Recently, a dedicated neural network was proposed to cope with current estimation during INS/DVL fusion \cite{Liu et al.2022}. 
\newline
The DVL emits four acoustic beams to the seafloor and measures the changes in the reflected beams' frequency. Using the frequency shift, the beam's velocity is calculated. The AUV velocity vector can be estimated when at least three beams are reflected back. In real-life scenarios, however, beams may not reflect back to the DVL for several reasons, such as if the AUV passes over a deep trench in one of the directions, an underwater sand wave changes the seafloor surface, or when the AUV operates in extreme roll and pitch angles. In such scenarios, the DVL cannot estimate the AUV velocity vector, and the INS/DVL loosely coupled approach cannot be applied. Since the tightly coupled approach uses any of the available beams, it can be implemented for the fusion process. Yet, for practical considerations, the loosely coupled method is usually implemented  \cite{Liu et al.2018,Yonggang et al. 2013}. To cope with situations of partial beam measurement, a model-based extended loosely coupled approach was suggested \cite{Tal et al.2017}.
\newline
The use of data-driven approaches in navigation and their benefits over model-based approaches were recently summarized in \cite{Klein 2022}. A novel method of improving the accuracy of the estimated DVL velocity in underwater navigation using a neural network structure was suggested \cite{Cohen and Klein. 2022}. Furthermore, a deep learning network that utilizes attitude and heading data in order to improve navigation accuracy and fault tolerance was developed  \cite{Zhang et al. 2020}.
\newline
This paper presents, a learning framework, MissBeamNet, to regress the missing DVL beams and enable AUV velocity vector estimation. To that end, we leveraged our initial research to regress only a single beam \cite{Yona and Klein 2021}.  The contributions of this research are:
\begin{itemize}
    \item A modular framework capable of regressing one, two, or three missing beams.
    \item  A robust long short-term memory network architecture able to accurately regress the missing beams.
    \item Inclusion of depth measurements to improve beam regression accuracy.
    \item A GitHub \href{https://github.com/ansfl/MissBeamNet}{repository} containing our code and dataset as a benchmark dataset and solution and to encourage further research in the field.
\end{itemize}
Here, we provide a thorough analysis of the missing beam scenarios. In addition, we compare our results to two model-based approaches: 1) an average of the missing beam to estimate the current one (baseline) and 2) the virtual beam approach \cite{Tal et al.2017}.  
All analyses were made on a dataset consisting of 11 hours of DVL recordings made by the Snapir AUV \cite{https://www.marinetech.haifa.ac.il/ocean-instruments} during its mission in the Mediterranean Sea. 
We further demonstrate the superiority of MissBeamNet over current model-based approaches and its ability to estimate the AUV velocity vector in situations of missing DVL beam measurements. 
\newline
The remainder of the paper is organized as follows: Section~\ref{sec:problem} describes the AUV sensors and the model-based partial DVL approaches. Section~\ref{sec:Learning Framework} presents our MissBeamNet framework, while Section~\ref{sec:Analysis Results} gives our sea experiment results. Finally, our conclusions are presented in Section~\ref{sec:Conclutions}.

\section{Problem Formulation}\label{sec:problem}
This section briefly describes the AUV sensors used in this research and presents the baseline model-based approaches to coping with missing beam measurements.
\subsection{AUV Sensors}
\subsubsection{DVL}\label{sec:DVL}
The DVL transmits four acoustic beams to the seafloor, which reach the seafloor and bounce back to the DVL transducers. The DVL measures the change in frequency in each direction. Based on \cite{Teledyne2008}, the relative velocity of each beam is calculated by:
\begin{equation}\label{eq:relative velocity}
	V_{rel} = (F_D +b_{F,D}+n_{F,D})\frac{1000\cdot C(1+SF_c)} {2f_s}
\end{equation}
where $F_D$ is the Doppler frequency shift, $b_{F,D}$ and $n_{F,D}$ are the bias and noise of the Doppler frequency shift, respectively, $SF_c$ is the scale factor error, $C$ is the speed of sound, and $f_s$ is the transmitted acoustic frequency.
The DVL transducers send acoustic beams in four directions. The standard DVL configuration is the "Janus Doppler configuration".In this configuration, the transducers are in an "X" shape, and the direction of each beam is described by the following equation:
\begin{equation}\label{eq:bvec}
	\mathbf{b_i} =
	\begin{bmatrix}
		\cos{\tilde{\psi}_i} \sin{\tilde{\theta}}   \\
		\sin{\tilde{\psi}_i} \sin{\tilde{\theta}}   \\
		\cos{\tilde{\theta}}
	\end{bmatrix}
\end{equation}
where $\tilde{\theta}$ is the (fixed) pitch angle and $\tilde{\psi}_i$ is the yaw angle defined for each beam $i$ as:
\begin{equation}
	\tilde{\psi}_i =(i-1)90\deg+45\deg,  i=1,2,3,4.
\end{equation}
The estimated DVL velocity in the platform frame is:
\begin{equation}\label{eq:vel_vce}
	\tilde{\mathbf{v}}^{p}_{t/p} = (\mathbf{A}^{T}\mathbf{A})^{-1}\mathbf{A}^{T}\mathbf{y}
\end{equation}
 where $\tilde{{v}}^{p}_{t/p}$ is the velocity vector,  $\mathbf{A}$ is the direction matrix defined as:
 \begin{equation}\label{eq:Amat}
	\mathbf{A} = \begin{bmatrix}
		\mathbf{b}^{T}_{1}
		\\
		\mathbf{b}^{T}_{2}
		\\
		\mathbf{b}^{T}_{3}
		\\
		\mathbf{b}^{T}_{4}
	\end{bmatrix}
\end{equation}
 and y is the measured beams vector
\begin{equation}\label{eq:dvl_all}
	\mathbf{y} = \left [ \begin{array}{cccc}
		\tilde{y}_{1} & \tilde{y}_{2} & \tilde{y}_{3} & \tilde{y}_{4}
	\end{array} \right ] ^{T}.
\end{equation}
\subsubsection{Pressure sensor}
A pressure sensor measures the pressure of a fluid or gas. In underwater navigation, a pressure sensor can be used to measure the water pressure at different depths, which can be used to determine the depth of the submerged vehicle. The underlying physical equation to estimate the AUV depth is \cite{Clayton et al.2016}:
\begin{equation}
    p= \rho \cdot g \cdot h+\rho p_0
\end{equation}
where $p$ [Kpa] is the measured pressure, $p_0$ is the pressure in the atmosphere equalling $101.3$[Kpa], $\rho$ is water density[$kg/m^3$], $g$ is the gravity magnitude, assumed here constant and equal to $9.81$ [$m/s^2$], and $h$ [m] is the depth of the AUV.\\
%

\subsection{Model-based approaches for missing beams}
\subsubsection{Average}\label{avg}
An average in a time window refers to the average value of a measurement over a specific period of time (the 'time window'). This can be useful for smoothing out noisy or erratic measurements, and reducing the effects of random errors. In the context of measurement synthesizing the average is a standard method that uses the average between the measurements in the previous time window, to assume the current measurement. For a time window with N measurements, the average is :
\begin{equation}
    AV(x) = \frac{1}{N}\sum^{N}_{k=1} x_k
\end{equation}
The size of the time window is chosen based on the characteristics of the sensor and system. For example, a small time window may be used for measurements that change rapidly, while a larger time window may be more suitable for relatively 
 stable measurements. 
\subsubsection{Virtual Beam}\label{VB}
The last velocity vector measurement can be utilized to predict the current velocity vector \cite{Tal et al.2017}. This method replaces the missing DVL beam measurement with the previously available measurement. For example, if beam \#1 is absent, solving  \eqref{eq:vel_vce} with the known velocity vector at $k-1$ gives an estimate of its velocity:
\begin{equation}
	y_{1,k} \approx \mathbf{b}^{T}_{1}[\hat{v}_{x,k-1} \quad \hat{v}_{y,k-1} \quad \hat{v}_{z,k-1}]
\end{equation}
where $k$ is the time index and $\hat{v}_{j,k-1}$ is the estimated velocity component from the previous step for $j=x,y,z$.
This approximated beam velocity is then used, together with the measured beams, in \eqref{eq:vel_vce} to predict the current velocity vector.
\section{MissBeamNet Framework}\label{sec:Learning Framework}
We propose a deep learning framework, MissBeamNet, as a mechanism to handle missing DVL beam measurements (1,2, or 3 beams) and allow the estimation of the AUV velocity vector. 
The MissBeamNet framework utilizes $n$ past DVL beam measurements and the currently available beams as input to an end-to-end neural network, which regresses the missing beams.
Then, the regressed and currently available measured beams are plugged into the model-based least squares (LS) estimator to estimate the AUV velocity vector. Figure 1 describes our MissBeamNet framework.\\
Our proposed MissBeamNet can cope with the following scenarios:
\begin{itemize}
    \item If three beams are available, MissBeamNet  will regress one missing beam.
    \item If two beams are available, MissBeamNet  will regress two missing beams.
    \item If one beam is available, MissBeamNet will regress three missing beams.
\end{itemize}
Note, that MissBeamNet was not designed to handle complete DVL outages, as it requires at least one available beam. For total outages, other solutions exist \cite{Lipman and klein. 2020,Lipman and klein. 2022}.
\begin{figure}[ht!] 
\centering
\includegraphics[width=3.0in]{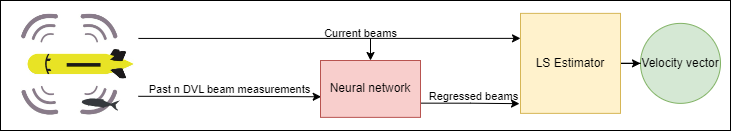}
	\caption{MissBeamNet framework utilizing past DVL beam measurements to regress the missing beams.}
	\label{fig:proposed approach}
\end{figure}
We consider two types of neural networks as our baseline network architectures. The first is based on a one-dimension  convolution neural network (CNN), while the other is based on long-short-term memory (LSTM) cells. Both networks have been proven to work with time-series data, such as those considered in our scenario. 
\subsection{Baseline Network Architectures }\label{sec:architectures}
\subsubsection{Convolutional Neural Network}\label{sec:CNN}
\begin{figure}[ht!] 
\centering
\includegraphics[width=3.0in]{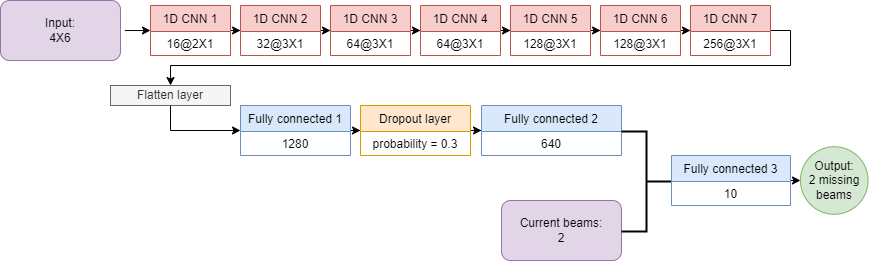}
	\caption{Baseline CNN architecture with an example of two missing beams.}
	\label{fig:1dcnn architecture}
\end{figure}

In CNN layers, there is a sparse interaction between the input and output, as appose to fully connected layers, where all the input parameters directly interact with the output. The convolution operator is a linear operator that involves multiplying an input with a kernel containing learned parameters. The kernel slides through the input, and the result is the sum of all the multiplications: 
\begin{equation}\label{eq:cnn eq}
    y_t = \sum_{k=1}^{p} x_{t+k} w_k
\end{equation}

where $t$ is the timestamp, $p$ is the kernel length, $w$ is the learned kernel parameter, and $x$,$y$ are the input and output, respectively. The fact that CNN shares parameters by passing the same kernels through all the input makes CNN architectures very popular in situations with large inputs. Figure \ref{fig:1dcnn architecture} describes our baseline CNN architecture, including network parameters, for a scenario of two missing beams. The network is a multi-head network where past DVL measurements are the input to the first head, and current DVL measurements are the input to the second head.
The same structure and parameters are used when one or three beams are missing. The selected activation function between the layers is Relu and the stride and padding are set to one.
\subsubsection{Long Short-Term Memory Network}\label{sec:LSTM}
\begin{figure}[ht!] 
\centering
\includegraphics[width=3.0in]{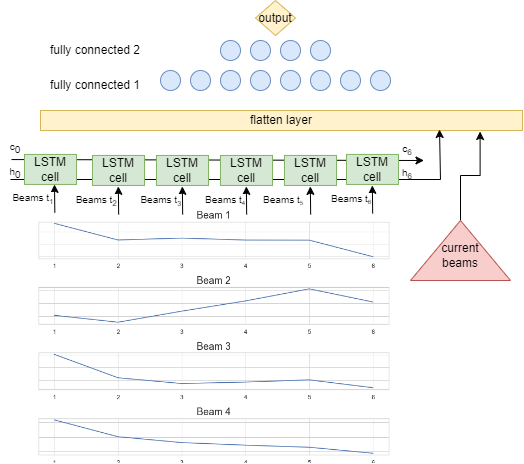}
	\caption{Baseline LSTM structure.}
	\label{fig:lstm fig}
\end{figure}

\begin{figure}[ht!] 
\centering
\includegraphics[width=3.0in]{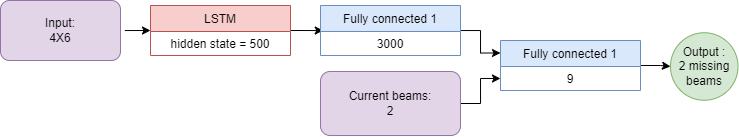}
	\caption{Baseline LSTM architecture with an example of two missing beams.}
	\label{fig:LSTM architecture parameters}
\end{figure}

LSTM is an advanced version of a recurrent neural network (RNN) and solves its shortcomings. RNNs are capable of handling temporal data by using information from prior inputs. 
However, if the sequence is long, the RNN may face a problem known as vanishing/exploding gradients \cite{Pascanu et al. 2013}. For example, when a gradient is small, it may continue to decrease until the model is no longer learning. The LSTM addresses these problems using three types of gates: The forget gate, the input gate, and the output gate.\\
The role of the forget gate is to forget unwanted information from the previous output and current input: 
\begin{equation}\label{eq:forget gate}
    f_t = \sigma (x_t U^f + h_{t-1}W^f +b_f)
\end{equation}
where $x_t$ is the input, $h_{t-1}$ is the output of the previous LSTM cell, $W^f$ and $b_f$ are the weights and biases of the forget gate, respectively. In \eqref{eq:forget gate} sigmoid function is employed to bring the parameter it wants to forget closer to zero. The output of the forget gate is then multiplied by the previous cell state.
The role of the input gate is to update the cell state 
$C_{t-1}$, by first calculating the input gate $i_t$:
\begin{equation}\label{input gate-sigmoid}
    i_t = \sigma (x_t U^i + h_{t-1}W^i+ b_i)
\end{equation}
where $U^i$ and $w^i$ are the gate weights and $b_i$ is the bias. 
Second, calculating the estimated cell state $\tilde{C_t}$:
\begin{equation}\label{input gate-tanh}
    \tilde{C_t} = tanh(x_t U^g + h_{t-1}W^g+b_g)
\end{equation}
where $U^g$ and $w^g$ are the gate weights and $b_g$ is the bias.
The results of \eqref{eq:forget gate},\eqref{input gate-sigmoid}, and \eqref{input gate-tanh} are used for the current cell state calculations:
\begin{equation}\label{eq:state}
    C_t = f_t \cdot C_{t-1} + i_t \cdot \tilde{C_t}
\end{equation}
As the name implies, the output gate $o_t$ determines which parameters are important as the output and next hidden state. 
\begin{equation}\label{output gate}
    o_t = \sigma (x_t U^o + h_{t-1}W^o+ b_o)
\end{equation}
where $U^t$ and $w^t$ are the gate weights and $b_t$ is the bias.
The output gate results are then multiplied by a tanh layer of the cell state to calculate the current output and hidden state
\begin{equation}\label{next hidden}
    h_t = o_t \cdot tanh(C_t)
\end{equation}
Figure \ref{fig:lstm fig} describes our LSTM baseline network structure. Previous beam measurements are used as input to the LSTM layers. After the LSTM features extraction, the features are concatenated with available beam measurements into a fully connected layer, which performs  the final process  resulting in the output of the regressed missing beams. Note that, like our baseline CNN network, this is a multi-head network where past DVL measurements are inputs to the first head, and current DVL measurements are inputs to the second head.
 Figure \ref{fig:LSTM architecture parameters} describes the LSTM architecture parameters in the scenario of two missing beams. The activation function between the layers is Relu. The same structure and parameters are also used when one or three beams are missing.
\subsection{Training Process}\label{sec:architectures}
The training process of deep neural networks requires defining a loss function. The common loss functions for regression problems are mean absolute error (MAE) or  mean squared error (MSE). In this paper, we use MSE loss defined by:
\begin{equation}\label{L2}
    MSE = \frac{1}{n} \sum_{i=1}^{n}(y_{actual} - y_{predicted})^2
\end{equation}
where $n$ is the number of samples, $y_{actual}$ is the target, and $y_{predicted}$ is the model output. Generally, the MSE loss function will try to adjust the model to better handle outliers than MAE due to the MSE squared error. However, in our scenarios, an AUV operates in varying sea conditions, therefore we adopt the MSE loss. During training, the loss function is calculated after each forward propagation in order to use the method of gradient descent and set the DNN initial weight and biases on values that will provide the desired result.
Forward propagation is the process of the data going through all the layers of the architecture, like \eqref{eq:cnn eq} for CNN and \eqref{eq:forget gate}-\eqref{next hidden} for LSTM networks. 
After completing the forward propagation process, the back propagation process updates the weights and biases of all the layers with a gradient descent principle
\begin{equation}\label{eq: gradient decent}
    \theta = \theta - \eta \nabla_\theta J(\theta)
\end{equation}
where $\theta$ is the vector of weights and biases, $J(\theta)$ is the loss function with the DNN weights and biases set to $\theta$, $\nabla_\theta$ is the gradient operator, and  $\eta$ is the learning rate.\\
The learning rate is a crucial hyperparameter, which dictates how fast the weights and biases change after each training batch. If the selected learning rate is too low, it might converge in a local minimum, and if it is too high, the model might not converge at a minimum. Our selected optimizer for all tested architectures is an adaptive moment estimation (ADAM) \cite{Diederik2014}.

\section{Analysis Results}\label{sec:Analysis Results}
\subsection{Dataset Description}\label{sec:data}
To examine the proposed approach, data from sea experiments were employed. All experiments were conducted in the Mediterranean Sea by the "Snapir" (ECA A18D), a 5.5[m] long AUV capable of reaching 3000[m] depth. It is equipped with the Teledyne RDI Work Horse navigator DVL\cite{teledynemarine_DVL}, which has a four-beams Janus convex configuration with a sample rate of $1$[Hz]. 
\begin{figure}[ht!] 
\centering
\includegraphics[width=3.3in]{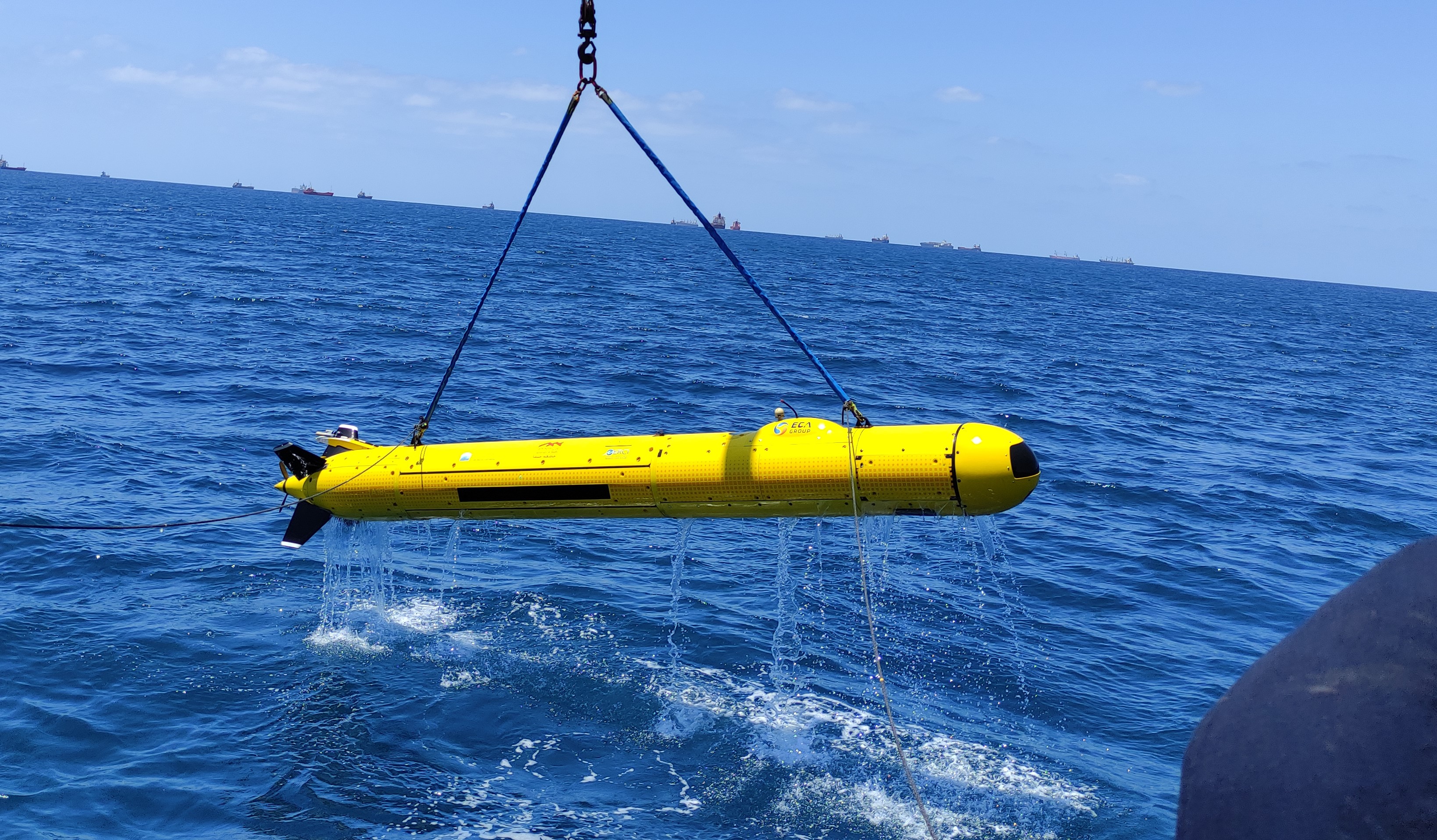}
	\caption{The "Snapir" being pulled out of the water after a successful mission.}
	\label{fig:auv pic}
\end{figure}
To train the deep neural network, first, all invalid DVL data was removed (some of the invalid readings occurred when actual beams 
 were missing). Then, the data was divided into routes that the AUV performed. Approximately 60\% of the missions were used as the training dataset and the rest as the test dataset. The training dataset comprised 23,243 samples corresponding to 387 minutes of recording, and the test dataset had 276 minutes of recording (16,618 samples).\\
Figure \ref{fig:route example} shows an experiment with challenging dynamics which is part of the test dataset.    
\begin{figure}[ht!] 
\centering
\includegraphics[width=3.3in]{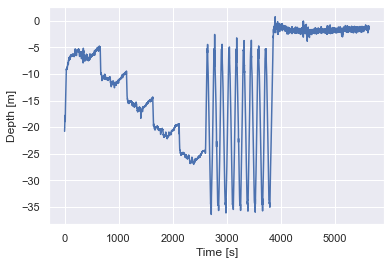}
	\caption{Experiment example from the test dataset.}
	\label{fig:route example}
\end{figure}

The total of 663 minutes of recordings consists of two parts:  300 minutes from our initial data collection campaign \cite{Shurin et al. 2022} and 363 minutes in the current campaign. The complete dataset is publicly available at our GitHub \href{repository}{https://github.com/ansfl/MissBeamNet}.
\subsection{Performance Metric}\label{sec:metric}
Performance metrics compare different models/methods and choose the one with the best performance. Throughout the research, we used the performance metric of root mean squared error (RMSE), which is widely used to evaluate models on regression tasks. RMSE is calculated by taking the root of the average of squared  differences between the predicted values and the target values
\begin{equation}\label{eq:RMSE}
    RMSE = \sqrt{\frac{\sum_{i=1}^{n}(y_{actual} - y_{predicted})^2}{n} }
\end{equation}
The RMSE results are in the same units as the original data, making it easy to interpret. 
\subsection{Baseline Architectures Comparison}\label{sec:stracture}
\begin{figure}[ht!] 
\centering
\includegraphics[width=3.3in]{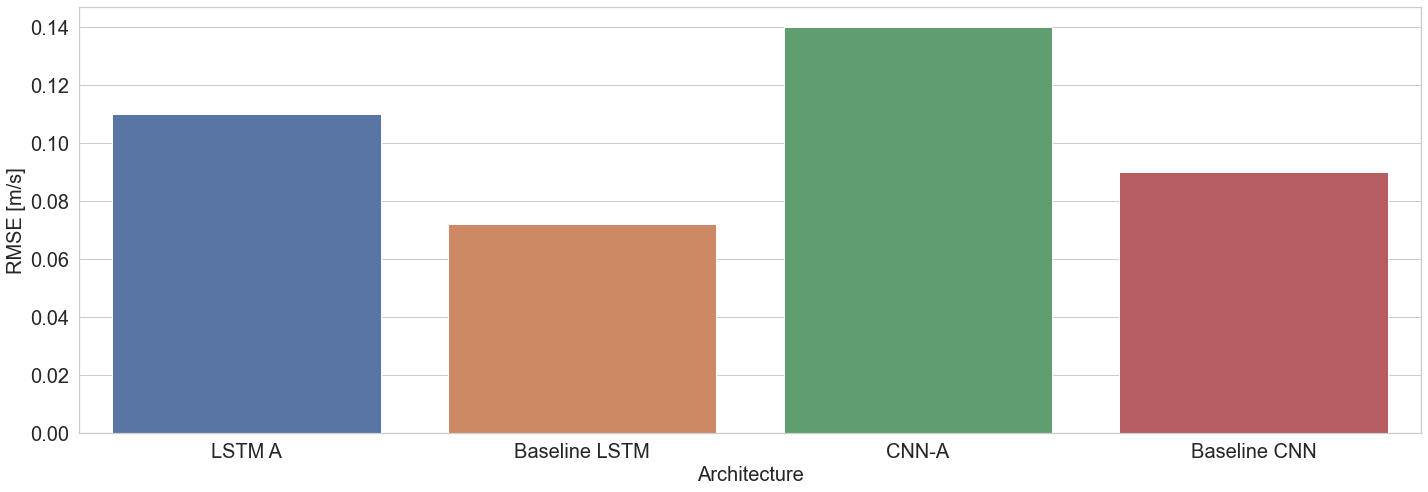}
	\caption{RMSE results for network architecture comparison.}
	\label{fig:architecure test}
\end{figure}

 To compare our two baseline architectures described in Section ~\ref{sec:architectures}, we consider a scenario with two missing beams, namely, beams \#1 and \#2, and assume six past beam measurements are used.  
In addition to these two baseline architectures, we examine the possibility of using only past beam measurements instead of the baseline multi-head approach. These two architectures are denoted as CNN A and LSTM A. The results of the test dataset in terms of RMSE are presented in Figure \ref{fig:architecure test}.
The results shows that using the baseline architectures (multi-head) obtained better performance than working with all the inputs in a single head.  In addition, the performance of the baseline LSTM showed an improvement of 27 \%  over the baseline CNN.\\

\subsection{Number of Past Beam Measurement Influence}\label{sec:windows-size}
\begin{figure}[ht!] 
\centering
\includegraphics[width=3.3in]{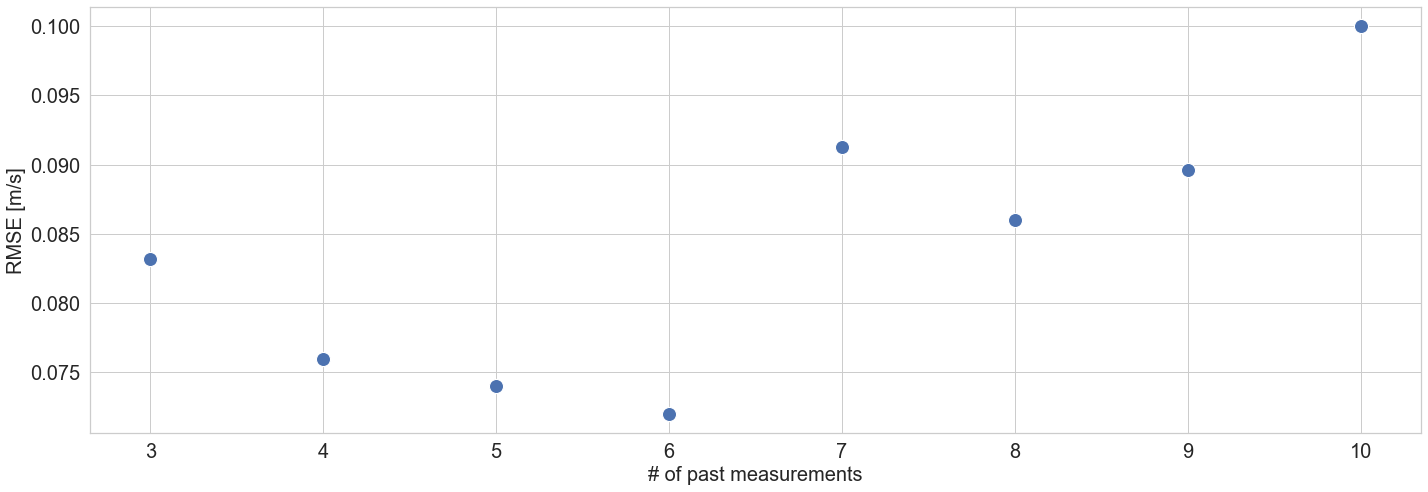}
	\caption{RMSE as a function of the window size for the baseline LSTM network.}
	\label{fig:window size}
\end{figure}

The number of past measurements utilized by the network is defined as the window-size length. The length of the optimal window size is crucial for model performance. The window size regularizes the model performance between the long and short movement patterns.  If the selected window size is too short, the model might miss the pattern of the AUV movement, and if it is too long, the model might not react well enough to a movement that just started. Figure \ref{fig:window size} shows the RMSE of the baseline LSTM model with different window-size lengths (between 3-10) when beams \#1 and \#2 are being regressed. 
 The results suggest that the optimal window-size length on our dataset is six measurements. \\
\subsection{Additional Input Information}\label{sec:additional data}
\begin{figure}[ht!] 
\centering
\includegraphics[width=3.3in]{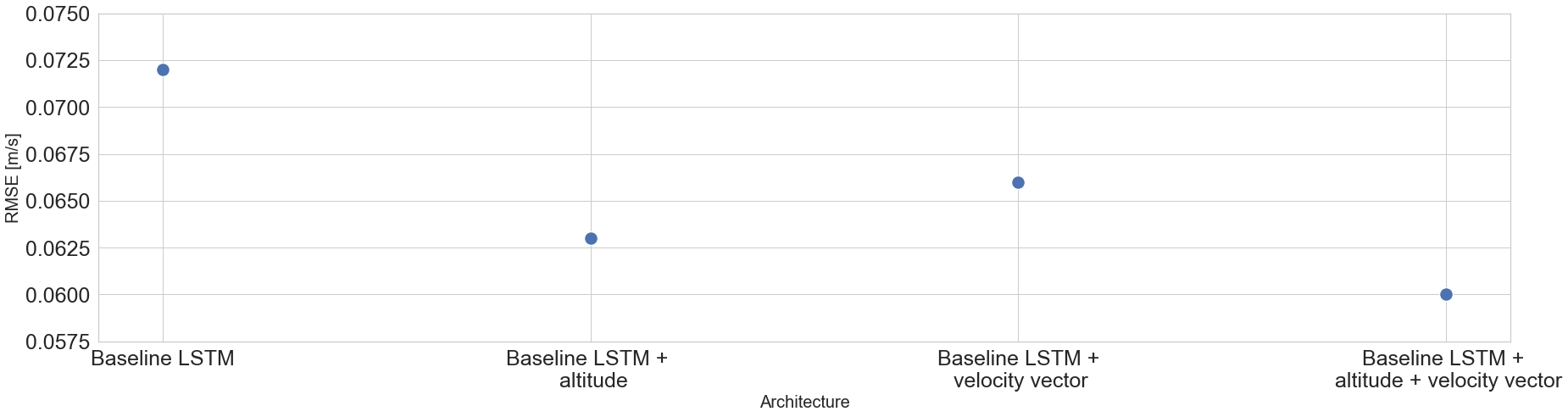}
	\caption{Velocity RMSE as a function of different input selection for our baseline LSTM network.}
	\label{fig:results additional}
\end{figure}

To improve the model performance even further, additional inputs are considered.
\begin{enumerate} 
 \item \textbf{Depth Sensor}: The last depth sensor reading.
 \item \textbf{AUV Velocity Vector}: Domain knowledge is used to transform the raw data (in this case, the beams) into meaningful features using feature engineering. 
Feature engineering is prevalent in classical machine learning methods, but less in deep neural networks. The assumption when using a neural network is thet model will learn the essential relations between features independently. The beams and the velocity vector are related, as the latter is estimated using the former. That is, the model is not receiving new information. Yet, in the proposed LSTM-based model, there are only two fully connected layers, and therefore feature engineering may help achieve better accuracy or shorten the network convergence time. 
\end{enumerate} 
Figure \ref{fig:results additional} describes the performance of each input with our baseline LSTM architecture, including additional inputs of 1) depth, 2) velocity vector, and 3) depth and velocity vector.  The tested case is when beams \#1 and \#2 are missing, and beams \#3 and \#4 are inserted as a two-phase input to our baseline LSTM network.
All of the additional inputs improved the performance of the baseline LSTM, and the best approach was obtained using all three input types - beam measurements (baseline), depth sensors, and the velocity vector. In this instance, there was a 16\% improvement compared to the LSTM baseline.

\subsection{Missing Beams Analysis}
There are 14 combinations of missing beams: four combinations of one missing beam, six of two missing beams, and four of three missing beams. In the proposed approach a different network needs to be trained for each of those combinations. Training for all networks used the same hyper-parameters: MSE loss function with a learning rate of 0.00005, batch size of 1 sequence, and 150 epochs.  In the following sections, we present the performance of our MissBeamNet approach compared to the average (baseline) and virtual beam approaches.  For this analysis, we employed our baseline LSTM network described in Section 3.1.2.  Based on the results of Section 4.4, we use six past DVL beam measurements and, based on Section 4.5, both the depth sensor reading and velocity vector were are added as additional inputs. The results were obtained on the testing dataset. 
\subsubsection{One missing beam} \label{sec: one missing beam res}
When one beam is missing, the least squared approach \eqref{eq:vel_vce} can be used to obtain the estimated AUV velocity vector. Table \ref{tab:one beam full data } presents the results of estimating the missing beam, the speed error obtained when using the estimated fourth beam together with the measured three, and the improvement of our MissBeamNet approach over the two model-based approaches.  

\begin{table*}
\begin{center}
\caption{One missing beam scenario results on the test dataset}
\scalebox{1}{
\begin{tabular}{c|c|c|c|c|c|c}
\hline
Case & Approach &Beam 1& Beam 2& Beam 3& Beam 4 & Avg. \\&&&&&&results \\
\hline

Missing beam [m/s] &Average (baseline) & 0.110&0.101 &0.101 &0.111 &0.106 \\

  &Virtual beam & 0.139&0.109 &0.110 &0.129 &0.121  \\

& MissBeamNet (ours) & 0.065& 0.048 & 0.034 &0.067 &0.053  \\
\hline

Speed error [m/s] & Average (baseline)& 0.066& 0.061&0.061 &0.067 &0.064  \\

 & Virtual beam & 0.079& 0.065&0.066 &0.077 &0.072  \\

& Three beams &0.450& 0.437&0.438 &0.449 &0.443  \\

& MissBeamNet (ours) & 0.039&0.029 &0.021 &0.040 &0.032 \\
\hline
 MissBeamNet improvement \% & Average (baseline)&40.9 & 52.4 &65.6 &40.3 & 49.8  \\

  &Virtual beam & 50.6 & 55.3 &68.1 &48.0 & 55.5 \\

&Three beams &91.3 & 93.3 &95.2 &91.1 &92.7  \\
\hline
\end{tabular}}
\label{tab:one beam full data }
\end{center}
\end{table*}
Both model-based and MissBeamNet methods were superior tp the three beams solution, indicating that regressing the fourth beam is critical to improving the AUV speed estimation accuracy. Specifically, MissBeamNet, improved the speed accuracy by over 90\%. In addition, MissBeamNet performed significantly better than the model-based approaches, with a 40\%-68\% improvement. Taking the mean of performance of all four scenarios  MissBeamNet improved the model-based approaches by over 49.8 \%.
\subsubsection{Two Missing Beams}
When considering two missing beams, six different combinations exist. In such scenarios, the AUV velocity cannot be estimated. Following the same procedure as the previous one missing beam scenarios, Table \ref{tab:two beams full} presents the results of two missing beams.
\begin{table*}[h]
	\caption{Two missing beam scenario results on the test dataset}
	\begin{center}\scalebox{1}
		{
			\begin{tabular}{c|c|c|c|c|c|c|c|c}
				\hline
				Case& Approach&\specialcell{Beam\\1,2}& \specialcell{Beam\\1,3}& \specialcell{Beam\\1,4}&\specialcell{Beam\\2,3} &\specialcell{Beam\\2,4} & \specialcell{Beam\\3,4} &\specialcell{Avg.\\result}  \\
				\hline
				Missing beams  [m/s] &Average (baseline) &0.106&0.106 &0.111 &0.101&0.107 &0.106 &0.106  \\ 
				
				&Virtual beam& 0.121&0.121 &0.131 &0.110&0.119 &0.120 &0.120 \\
				
				&MissBeamNet (ours) &0.062& 0.052 & 0.085& 0.076 & 0.057& 0.066 & 0.066  \\
				
				\hline
				Speed error [m/s] &Average (baseline)& 0.092& 0.077&0.096 & 0.088 &0.079& 0.092 &0.087 \\
				
				&Virtual beam &0.106& 0.069&0.114 & 0.096 &0.065& 0.105 &0.092 \\
				
				&MissBeamNet (ours) & 0.055 &0.057 &0.075 &0.066 &0.061 & 0.058 &0.062 \\
				
				\hline
				MissBeamNet improvement \% & Average (baseline) & 40.2 & 25.9 &21.9 & 25.0  & 22.8&36.9&28.7\\
				
				&Virtual beam & 48.1 & 17.4 &34.2 & 31.25  & 6.15&44.7 &30.3\\
				\hline

		\end{tabular}}
		\label{tab:two beams full}
	\end{center}
\end{table*}
The results show a significant difference between the speed error in each combination, even in the model-based approaches, emphasizing the problem's complexity. Yet, in all cases, MissBeamNet was more accurate than the model-based approaches, with a minimum improvement of 20\% that reached almost 50\%. The average improvement over the baseline model-based approach was 28.7\% compared to 49.8\% when only one beam was missing. This is attributed to the model receiving less information from two current beams compared to three when only one is missing.
\subsubsection{Three Missing Beams}
Table \ref{tab:three beams full} presents the results for the four scenarios in which three beams are missing.
\begin{table*}[h]
	\caption{Three missing beam scenario results on the test dataset}
	\begin{center}
			\begin{tabular}{c|c|c|c|c|c|c}
				\hline
				Case & Approach & \specialcell{Beam \\1,2,3}& \specialcell{Beam \\2,3,4}&\specialcell{Beam \\1,2,4} &\specialcell{Beam \\1,3,4} &\specialcell{Avg.\\results}\\
				\hline
				
				Missing beams [m/s]  &Average (baseline)& 0.104 &0.108&0.107 &0.105 &0.106 \\
				
				 &Virtual beam & 0.118 &0.124 &0.124 &0.116 &0.120 \\
				
				&MissBeamNet (ours) & 0.071& 0.073 & 0.077& 0.071&0.073  \\
				\hline
				Speed error [m/s] & Average (baseline) & 0.102 & 0.108 &0.106& 0.103 &0.105 \\
				
				&Virtual beam & 0.102 & 0.109 &0.111& 0.099&0.105 \\
				
				&MissBeamNet (ours)  &0.077 &0.081 &0.083 & 0.078 &0.079  \\
						
				\hline
				MissBeamNet improvement \% &Average (baseline) & 24.5 & 25.0&21.7 &24.3 &23.9 \\
				
				&Virtual beam & 24.5 & 25.7&25.2 &21.2 &24.1 \\
				\hline
		\end{tabular}
		\label{tab:three beams full}
	\end{center}
\end{table*}
As expected, the speed error when three beams are missing is higher than in the two or one missing beams scenarios. Yet, MissBeamNet use improved results by at least 21\% over the model-based approaches. For three missing beams, the average improvement was 24\% compared to 28.7\% when two beams were missing, only 4.7\% less, indicating that even with only one beam at hand, the AUV velocity can be estimated. 
\subsubsection{Hyperparameter Tuning}
One of the main challenges in deep learning research is to find the best combination of hyperparameters for the proposed architecture. 
Each architecture has several parameters that can influence model performance, including the number of layers, the number of parameters in each layer, the type of cost function, the learning rate, and batch size. To demonstrate the potential of hyperparameter tuning, we evaluated three different hyperparameters.  The first was the learning rate, which affects how much each batch changes the weights and biases \ref{sec:architectures}. The second hyper parameter was the number of hidden parameters in the LSTM layer $h_{t}$ \ref{sec:LSTM}, and the third  hyperparameter is the number of parameters in the LSTM output. To test the importance of hyperparameter tuning, each parameter was set with a few available options, and a seed was set (equal initialization in each run).
We focused on a one missing beam scenario, which has four options - missing beam $\#1, \#2, \#3$, or $\#4$. For each case, 15 randomly selected combinations of the three hyperparameters were examined. 
\begin{table}[h]
	\caption{hyperparameters tuning}
	\begin{center}
		\scalebox{0.8}{
			\begin{tabular}{c|c|c|c|c|c}\label{tab:hyperparameters}

				Case& Approach &Beam 1& Beam 2& Beam 3& Beam 4 \\
				\hline
				Missing beam [m/s] &Before tuning & 0.065& 0.048 & 0.034 &0.067  \\
				
				& After tuning & 0.011&0.017 &0.02 &0.012  \\
				
                \hline

				Speed error [m/s] & Before tuning & 0.039&0.029 &0.021 &0.040 \\
				
				& After tuning & 0.007&0.010 &0.012 &0.007 \\
				\hline

				Learning rate & Before tuning & 5e-05& 5e-05 & 5e-05 &5e-05 \\
				
				& After tuning & 1e-04& 1e-04 &5e-05 &1e-05 \\
				\hline

				\specialcell{hidden LSTM \\parameters $h_{t}$ } & Before tuning & 500& 500 & 500 &500 \\
				
				& After tuning & 250& 750 &750 &100 \\
				\hline

				 \specialcell{LSTM output\\ parameters} & Before tuning & 7& 7 & 7 &7 \\
				
				& After tuning & 7& 5 &7 &5 \\
                \hline
                
				 \specialcell{MissBeamNet Tuning \\improvement \%} & Average (baseline) &89.4 & 83.6 &80.3 &89.5  \\
                        				 & Virtual beam &91.1 & 84.6 &81.8 &90.9  \\
                        				 & Three beams &98.4 & 97.7&97.2 &98.4 \\
                        				 & Before tuning &82.2 & 67.5 &42.8 &82.5  \\
				
				\hline
		\end{tabular}}
	\end{center}
\end{table}
Table \ref{tab:hyperparameters} presents the potential of hyperparameter tuning. It is important to note that out of the 15 tested hyperparameter combinations, only a few were better than the results before tuning. Yet, they were able to improve the missing beam estimation and, consequently, reduce the speed error and increase the rate of improvement compared to the two model-based approaches. 
\section{Conclusions}\label{sec:Conclutions}
Here, we presented MissBeamNet, a deep learning-based framework developed to compensate for partial DVL measurement scenarios (1, 2, or 3 missing beams). To that end, an LSTM-based dedicated DNN was derived. We demonstrated that the best input to the network is past DVL measurements, past depth sensor measurements, previous velocity vectors, and the currently available measured beams. Once the missing beams are regressed, they are combined with the available beams and plugged into the classical model-based approach to estimate  the AUV velocity vector.\\
To evaluate MissBeamNet, sea experiments with the University of Haifa’s "Snapir" AUV were conducted. The data included several trajectories collected for different purposes and under various sea conditions.
We provide a thorough analysis of all 14 missing beam combinations and explore several means to enhance our baseline architecture. The results show that MissBeamNet allows estimating the missing DVL beams and, consequently, the AUV velocity vector. Additionally, MissBeamNet significantly improves the accuracy of the velocity vector in all examined scenarios compared to the model-based approaches. The improvement of all three missing beam combinations was above 20 \% over the model-based approaches. For two missing beams, performance was generally better compared to three missing beams since the model uses one additional measured beam. Finally, we show that hyperparameters-tuned models improve the accuracy of MissBeamNet by more than $40\%$.

\bibliographystyle{IEEEtran}

\end{document}